\documentclass[10pt,twocolumn,letterpaper]{article}

\usepackage{iccv}
\usepackage{times}
\usepackage{epsfig}
\usepackage{graphicx}
\usepackage{amsmath}
\usepackage{amssymb}

\usepackage[breaklinks=true,bookmarks=false]{hyperref}

\iccvfinalcopy 

\def\iccvPaperID{2423} 

\ificcvfinal\fi
\begin{document}

\title{Predicting Deep Zero-Shot Convolutional Neural Networks\\using Textual Descriptions}

\author{Jimmy Lei Ba
\qquad
Kevin Swersky
\qquad
Sanja Fidler 
\qquad
Ruslan Salakhutdinov \\
University of Toronto\\
{\tt\small jimmy,kswersky,fidler,rsalakhu@cs.toronto.edu}
}

\maketitle

\begin{abstract}
  One of the main challenges in Zero-Shot Learning of visual categories is 
  gathering semantic attributes to accompany images. Recent work has shown that
  learning from textual descriptions, such as Wikipedia articles, avoids the
  problem of having to explicitly define these attributes. We present a new
  model that can classify unseen categories from their textual description.
  Specifically, we use text features to predict the output weights of both the
  convolutional and the fully connected layers in a deep convolutional neural
  network (CNN). We take advantage of the architecture of CNNs and learn
  features at different layers, rather than just learning an embedding space
  for both modalities, as is common with existing approaches. The proposed
      model also allows us to automatically generate a list of
      pseudo-attributes for each visual category consisting of words from
      Wikipedia articles. We train our models end-to-end using the Caltech-UCSD
      bird and flower datasets and evaluate both ROC and Precision-Recall
      curves.  Our empirical results show that the proposed model significantly
      outperforms previous methods. 
\end{abstract}

\section{Introduction}

The recent success of the deep learning approaches to object recognition is supported by the collection of large datasets with millions of images and thousands of labels~\cite{imagenet,ZhouNIPS2014}. Although the datasets continue to grow larger and are acquiring a broader set of categories, they are very time consuming and expensive to collect. Furthermore, collecting detailed, fine-grained annotations, such as attribute or object part labels, is even more difficult for datasets of such size.


On the other hand, there is a massive amount of textual data available online. Online encyclopedias, such as English Wikipedia, currently contain 4,856,149 articles, and represent a rich knowledge base for a diverse set of topics.  Ideally, one would exploit this rich source of information in order to train visual object models with minimal additional annotation. 
\begin{figure}
\includegraphics[scale=0.235]{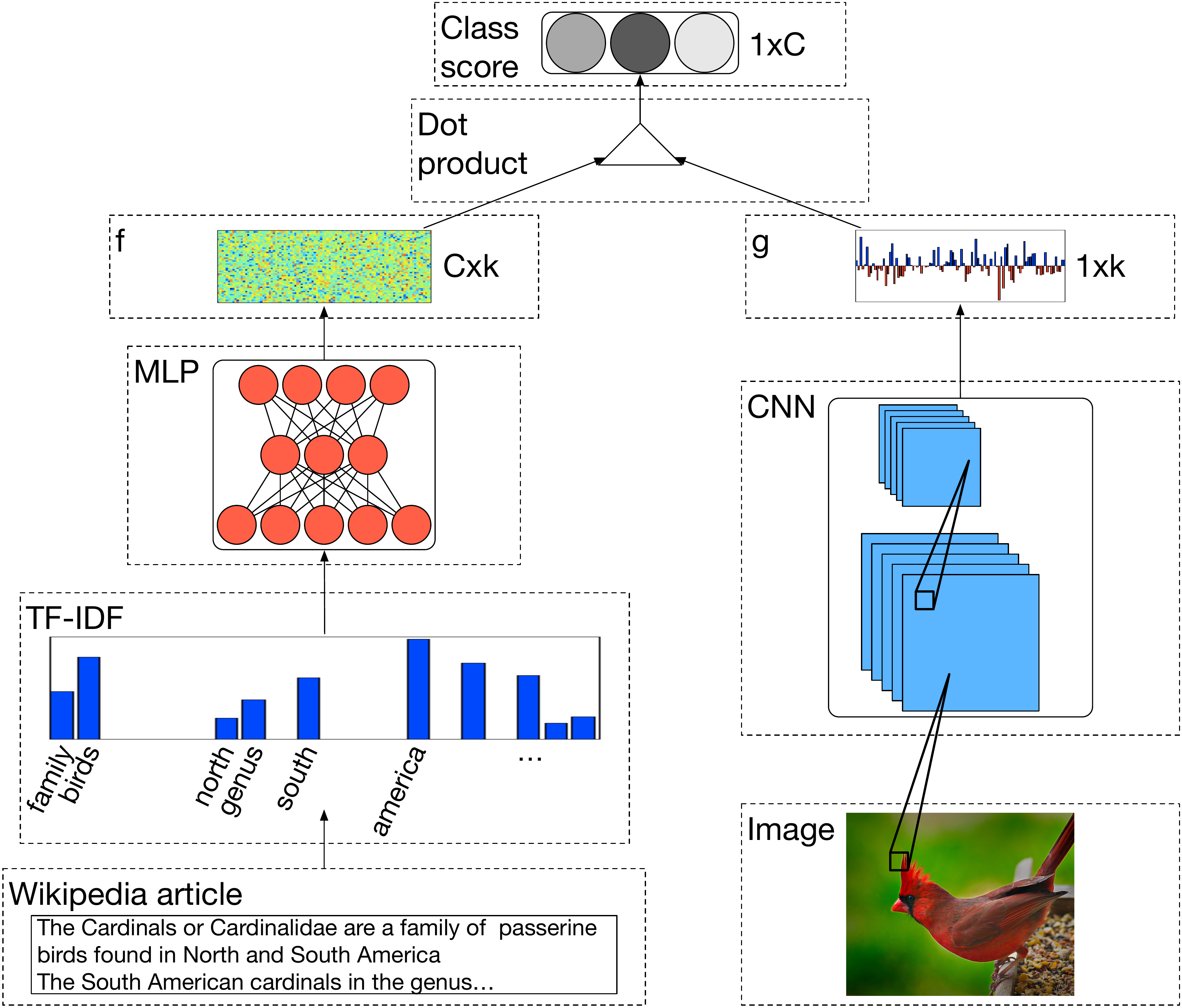}
\label{fig:model}

\caption{{A deep multi-modal neural network. The first modality corresponds to tf-idf features taken from a text corpus with a corresponding class, e.g., a Wikipedia article about a particular object. This is passed through a multi-layer perceptron (MLP) and produces a set of linear output nodes $f$. The second modality takes in an image and feeds it into a convolutional neural network (CNN). The last layer of the CNN is then passed through a linear projection to produce a set of image features $g$. The score of the class is produced via $f^\top g$. In this sense, the text pipeline can be though of as producing a set of classifier weights for the image pipeline.}}
\end{figure}

 The concept of ``Zero-Shot Learning'' has been introduced in the literature~\cite{feifef03,fink04,lampert09,Palatucci09,writeaclassifier,larochelle2008zero} with the aim to improve the scalability of traditional object recognition systems. The ability to classify images of an unseen class is transferred from the semantically or visually similar classes that have already been learned by a visual classifier. One popular approach is to exploit shared knowledge between classes in the form of attributes, such as \emph{stripes}, \emph{four legs}, or \emph{roundness}. 
 There is typically a much smaller perceptual (describable) set of attributes than the number of all objects, and thus training classifiers for them is typically a much easier task. Most work pre-defines the attribute set, typically depending on the dataset used, which somewhat limits the applicability of these methods on a larger scale.

In this work, we build on the ideas of \cite{writeaclassifier} and introduce a novel Zero-Shot Learning model that predicts visual classes using a text corpus, in particular, the encyclopedia corpus. The encyclopedia articles are an explicit categorization of human knowledge. Each article contains a rich implicit annotation of an object category. For example, the Wikipedia entry for ``Cardinal''  gives a detailed description about this bird's distinctive visual features, such as colors and shape of the beak. The explicit knowledge sharing in encyclopedia articles are also apparent through their inter-references. Our model aims to generate image classifiers directly from encyclopedia articles of the classes with no training images. This overcomes the difficulty of hand-crafted attributes and the lack of fine-grained annotation. Instead of using simple word embeddings or short image captions, our model operates directly on a raw natural language corpus and image pixels.

Our first contribution is a novel framework for predicting the output weights of a classifier on both the fully connected and convolutional layers of a Convolutional Neural Network (CNN). We introduce a convolutional classifier that operates directly on the intermediate feature maps of a CNN. The convolutional classifier convolves the feature map with a filter predicted by the text description. The classification score is generated by global pooling after convolution. We also empirically explore combining features from different layers of CNNs and their effects on the classification performance.

We evaluate the common objective functions used in Zero-Shot Learning and rank-based retrieval tasks. We quantitatively compare performance of different objective functions using ROC-AUC, mean Average-Precision and classification accuracy. We show that different cost functions outperform each other under different evaluation metrics. Evaluated on Caltech-UCSD Bird dataset and Oxford flower dataset, our proposed model significantly outperforms the previous state-of-the-art Zero-Shot Learning approach~\cite{writeaclassifier}. In addition, the testing performance of our model on the seen classes are comparable to the state-of-the-art fine-grained classifier using additional annotations.  
 
Finally, we show how our trained model can be used to automatically discover a list of class-specific attributes from encyclopedia articles.  

\section{Related work}

\subsection{Domain adaptation}
Domain adaptation concerns the problem where there are two distinct datasets, known as the \emph{source} and \emph{target} domains respectively. In the typical supervised setting, one is given a source training set $\mathcal{S} \sim P_\mathcal{S}$ and a target training set $\mathcal{T} \sim P_\mathcal{T}$, where $P_\mathcal{S} \neq P_\mathcal{T}$. The goal is to transfer information from the source domain to the target domain in order to produce a better predictor than training on the target domain alone. Unlike zero-shot learning, the class labels in domain adaptation are assumed to be known in advance and fixed.

There has been substantial work in computer vision to deal with domain adaption. \cite{quionero2009dataset, saenko2010adapting} address the problem mentioned above where access to both source and target data are available at training time. This is extended in \cite{gopalan2011domain} to the unsupervised setting where target labels are not available at training time. In \cite{torralba2011unbiased}, there is no target data available, however, the set of labels is still given and is consistent across domains. In~\cite{khosla2012undoing} the authors explicitly account for inter-dataset biases and are able to train a model that is invariant to these. \cite{yang2014unified} considered unified formulation of domain adaptation and multi-task learning where they combine different domains using a dot-product operator.

\subsection{Semantic label embedding}
Image and text embeddings are projections from the space of pixels, or the space of text, to a new space where nearest neighbours are semantically related. In semantic label embedding, image and label embeddings are jointly trained so that semantic information is shared between modalities. For example, an image of a tiger could be embedded in a space where it is near the label ``tiger'', while the label ``tiger'' would itself be near the label ``lion''.

In \cite{weston2011wsabie}, this is accomplished via a ranking objective using linear projections of image features and bag-of-words attribute features. In \cite{frome2013devise}, label features are produced by an unsupervised skip-gram model~\cite{mikolov2013distributed} trained on Wikipedia articles, while the image features are produced by a CNN trained on Imagenet~\cite{alexnet}. This allows the model to use semantic relationships between labels in order to predict labels that do not appear in the training set. While \cite{frome2013devise} removes the final classification layer of the CNN, \cite{norouzi2014iclr} retains it and uses the uncertainty in the classifier to produce a final embedding from a convex combination of label embeddings. \cite{socher2013zero} 
uses unsupervised label embeddings together with an outlier detector to determine whether a given image corresponds to a known label or a new label. This allows them to use a standard classifier when the label is known.

\subsection{Zero-Shot learning from attribute vectors}
A key difference between semantic label embedding and the problem we consider here is that we do not consider the semantic relationship between labels. Rather, we assume that the labels are themselves composed of attributes and attempt to learn the semantic relationship between the attributes and images. In this way, new labels can be constructed by combining different sets of attributes. This setup has been previously considered in \cite{farhadi2009describing, lampert09}, where the attributes are manually annotated. In \cite{farhadi2009describing}, the training set attributes are predicted along with the image label at test time. \cite{parikh2011relative} explores relative attributes, which captures how images relate to each other along different attributes.

Our problem formulation is inspired by \cite{writeaclassifier} in that we attempt to derive embedding features for each label directly from natural language descriptions, rather than attribute annotations. The key difference is in our architecture, where we use deep neural networks to jointly embed image and text features rather than using probabilistic regression with domain adaptation.

\section{Predicting a classifier}

The overall goal of the model is to learn an image classifier from
natural language descriptions. During training, our model takes a set of text features (e.g. Wikipedia articles), 
each representing a particular class, and a set of images for each class. During test time, some previously unseen textual description (zero-shot classes) and associated images are presented. Our model needs to classify the images from unseen visual classes against images from the trained classes. We first introduce a general framework to predict linear classifier weights and extend the concept to convolutional classifiers.

Given a set of $N$ image feature vectors $x \in R^d$ and their associated class
labels $l \in \{1, ..., C\}$, we have a training set $\mathcal{D}_{train} = \{(x^{(n)}, l^{(n)})\}_N$. There are $C$ distinct class labels available for training. During test time, we are given additional $n_0$ number of the previously unseen classes, such that $l_{test} \in \{1, ..., C,...C+n_0\}$ and test images $x_{test}$ associated with those unseen classes, $\mathcal{D}_{test} = \{(x^{(n)}_{test}, l^{(n)}_{test})\}_{N_{test}}$.

\subsection{Predicting a linear classifier}
Let us consider a standard binary one vs.
all linear classifier whose score is given by\footnote{We consider various loss functions of this score in Section \ref{sec:learning}.}:
\begin{align}
\label{eq:ft}
\hat{y}_c=w_c^{\top}x,
\end{align}
where $w_c$ is the weight vector for a particular class $c$. 
It is hard to deal with unseen classes using this standard formulation. Let us further 
assume that we are provided with an additional text feature vector $t_c \in \mathbb{R}^p$ associated 
with each class $c$.
Instead of learning a static weight vector $w_c$, the text feature can be used
to predict the classifier weights $w_c$. In the other words, we can define $w_c$ to be a
function of $t_c$ for a particular class $c$:
\begin{align}
w_c=f_t(t_c),
\end{align}
where, $f_t:\mathbb{R}^p \mapsto \mathbb{R}^d$ is a mapping that transforms the text features to the visual image feature space. 
In the special case of choosing $f_t(\cdot)$ to be a linear transformation, the formulation is similar to \cite{whatyousee}. In this work, the mapping $f_t$ is represented as a non-linear regression model that is 
parameterized by a neural network. Given the mapping $f_t$ and text features for a new class, we can extended the one-vs-all linear classifier to the previously unseen classes.  

\subsection{Predicting the output weights of neural nets}
\label{sec:predfc}
One of the drawbacks for having a direct mapping from $\mathbb{R}^p$ to $\mathbb{R}^d$  is that both $\mathbb{R}^p$ and $\mathbb{R}^d$ are typically high dimensional, which 
makes it difficult to estimate the large number of parameters in $f_t(\cdot)$. For example, 
in the linear transformation setup, the number of parameters in $f_t(\cdot)$ is proportional to $O(d\times p)$. 
For the problems considered in the paper, this implies that millions of parameters need to be estimated from only a few thousand data points. In addition, most the parameters are highly correlated which makes gradient based optimization methods converge slowly. 

Instead, we introduce a second mapping parameterized by a multi-layer neural network 
$g_v:\mathbb{R}^d \mapsto \mathbb{R}^k$ that transforms the visual image features $x$ to a lower dimensional space
 $\mathbb{R}^k$, where $k<<d$. The dimensionality of the predicted weight vector $w_c$ can be drastically reduced using $g_v(\cdot)$. The new formulation for the binary classifier can be written as:
 \begin{align}
\label{eq:gv}
\hat{y}_c=w_c^{\top}g_v(x),
\end{align}
where the transformed image feature $g_v(x)$ is the output of a neural network. Similar to Eq. (\ref{eq:ft}), $w_c \in \mathbb{R}^k$ is predicted using the text features $t_c$ with $f_t: \mathbb{R}^p \mapsto \mathbb{R}^k$. Therefore, the formulation in the Eq. (\ref{eq:gv}) is equivalent to a binary classification neural network whose output weights are predicted from text features. Using neural networks, both $f_t(\cdot)$ and $g_t(\cdot)$ perform non-linear dimensionality reduction of the text and visual features. In the special case 
where both $f(\cdot)$ and $g(\cdot)$ are linear transformations, Eq. (\ref{eq:gv}) is equivalent to the low rank matrix factorization~\cite{whatyousee}. A visualization of this model is shown in Figure \ref{fig:model}.

\subsection{Predicting a convolutional classifier}
\label{sec:predconv}
Convolutional neural networks (CNNs) are currently the most accurate models for object recognition tasks~\cite{alexnet}. In contrast to traditional hand-engineered features, CNNs build a deep hierarchical multi-layer feature representation from raw image pixels. It is common to boost the performance of a vision system by using the features from the fully connected layer of a CNN~\cite{donahue2013decaf}.  Although, the image features obtained from the top fully connected layer of CNNs are useful for generic vision pipelines, there is very little spatial and local information retained in them. The feature maps from the lower convolutional layers on the other hand contain local features arranged in a spatially coherent grid. In addition, the weights in a convolution layer are locally connected and shared across the feature map. The number of trainable weights are far fewer than the fully connected layers. It is therefore appealing to predict the convolutional filters using text features due to the relatively small number of parameters. 

Let $a$ denote the extracted activations from a convolutional layer with $M$ feature maps,
where $a \in \mathbb{R}^{M\times w \times h}$ with 
$a_i$ representing the $i^{th}$ feature map of $a$, and $w$, $h$ denoting the width and height of a feature map. 
Unlike previous approaches, we directly formulate a convolutional classifier using the feature maps from convolutional layers. First, we perform a non-linear dimensionality reduction to reduce the 
number of feature maps as in Sec. (\ref{sec:predfc}). Let $g'_v(\cdot)$ be a reduction mapping $g'_v: \mathbb{R}^{M\times w\times h} \mapsto \mathbb{R}^{K'\times w\times h}$ where $K' << M$. 
The reduced feature map is then defined as $a' = g'_v(a)$. Given the text features $t_c$ for a 
particular class $c$, we have the corresponding predicted convolutional weights $w'_c = f'_t(t_c)$, where $w'_c \in \mathbb{R}^{K' \times s \times s}$ and $s$ is the size of the predicted filter. Similarly to the fully connected model, $f'_t(\cdot)$ is parameterized by a multi-layer neural network.  
We can formulate a convolutional classifier as follows:
\begin{align}
\label{eq:conv}
\hat{y'}_c=o\bigg(\sum_{i=1}^{K'}w'_{c,i}\  \check{*} \  a'_i\bigg),
\end{align}
where $o(\cdot)$ is a global pooling function such that $o:\mathbb{R}^{w\times h} \mapsto \mathbb{R}$ and $\check{*}$ denotes the convolution that is typically used in convolutional layers. 
By convolving the predicted weights over the feature maps, we encourage the model to learn informative location feature detectors based on textual descriptions. The global pooling $o(\cdot)$ operation aggregates the local features over the whole image and produces the score. Depending on the type of the pooling operation, such as noisy-or average pooling or max pooling, the convolutional classifier will have different sensitivities to local features. In our experimental results, we found that average pooling works well in general while max pooling suffers from over-fitting. 

\subsection{Predicting a joint classifier}

We can also take advantage of the CNN architecture by using features extracted from both the intermediate convolutional layers and the final fully connected layer. Given convolutional feature $a$ and fully connected feature $x$ after propagating the raw image through the CNN, we can write down the joint classification model as:
\begin{align}
\hat{y}_c=w_c^Tg_v(x) + o\bigg(\sum_{i=1}^{K'}w'_{c,i}\  \check{*} \  g'_v(a)_i\bigg).
\end{align}
Both the convolutional weights $w'_c$ and the fully connected weights $w_c$ are predicted from the text feature $t_c$ using a single multi-task neural network with shared layers.

\section{Learning}
\label{sec:learning}
The mapping functions $f(\cdot)$ and $g(\cdot)$ that transform text features into weights are neural 
networks that are parameterized by a matrix $W$. The goal of learning is to adjust $W$ so that the model 
can accurately classify images based on a textual description. 
Let us consider a training set containing $C$ textual descriptions (e.g. $C$ Wikipedia articles), one for each class $c$, 
and $N$ images. 
We next examine the following two objective functions for training our model.

\subsection{Binary Cross Entropy}
\label{sec:bce}
For an image feature $x_i$ and a text feature $t_j$, an indicator $I_{ij}$ is used to encode whether the image corresponds to the class represented by the text using a 0-1 encoding. The binary cross entropy is the most intuitive objective function for our predicted binary classifier:
\begin{align}
\mathcal{L}(W)=& \sum_{i=1}^N\sum_{j=1}^C \bigg[ I_{i,j}\log \sigma(\hat{y}_j(x_i, t_j)) \nonumber \\
        &+ (1-I_{i,j})\log (1-\sigma(\hat{y}_j(x_i, t_j))) \bigg],
\label{eq:bce}
\end{align}
{where $\sigma$ is the sigmoid function $y=1/(1+e^{-x})$.} In the above equation, each image is evaluated against all $C$ classes during training, 
which becomes computationaly expensive as the number of classes grows. 
Instead, we use a Monte Carlo minibatch scheme 
to approximate the summation over the all images and all classes from Eq. (\ref{eq:bce}). 
Namely, we draw a mini-batch of $B$ images and compute the cost by summing over the images in the minibatch.
We also sum over all the image labels from the minibatch only. The computational cost for this 
minibatch scheme is $O(B\times B)$, instead of $O(N \times C)$.

\subsection{Hinge Loss}
\label{sec:hinge}
We further considered a hinge loss objective. 
Hinge loss objective functions are the most popular among the retrieval and ranking tasks for multi-modal data. 
In fact, predicting the output layer weights of a neural network  (see Sec. (\ref{sec:predfc})) 
can be formulated as a ranking task between text descriptions and visual images. Although the formulation is similar, the focus of this work is on classification rather than information retrieval. 
Let the indicator $I_{i,j}$ represent a $\{1,-1\}$ encoding for the positive and negative class. 
We can then use the following simple hinge loss objective function:
\begin{align}
\mathcal{L}(W)=& \sum_{i=1}^N\sum_{j=1}^C \max (0, \epsilon -  I_{i,j} \hat{y}_j(x_i, t_j)).
\label{eq:hinge}
\end{align}
Here, $\epsilon$ is the margin that is typically set to 1. This hinge loss objective encourages the classifier score $\hat{y}$ to be higher for the correct text description and lower for other classes. Similarly to Sec. (\ref{sec:bce}), a minibatch method can be adapted to train the hinge loss objective function efficiently.

\subsubsection{Euclidean Distance}
\label{sec:l2dis}
The Euclidean distance loss function was previously used in \cite{socher2013zero} with a fixed pre-learnt word embedding. Such cost function can be obtained from our classifier formulation by expanding the Euclidean distance $-{1\over2}\| a - b \|^2_2 = a^Tb -{1\over2}\|a\|^2_2  -{1\over2}\|b\|^2_2 $. Minimizing the hinge loss in Eq. (\ref{eq:hinge}) with the additional negative $L_2$ norm of both $w_c$ and $g_v$ is equivalent to minimizing their Euclidean distance. The hinge loss prevents the infinite penalty on the negative examples when jointly learning an embedding of class text descriptions and their images.

\section{Experiments}

\newcommand{\ts}{\rule{0pt}{2.6ex}}       
\newcommand{\ms}{\rule{0pt}{0ex}}         
\newcommand{\bs}{\rule[-1.2ex]{0pt}{0pt}} 
\newcommand{\specialcell}[2][c]{%
  \begin{tabular}[#1]{@{}c@{}}#2\end{tabular}}

\begin{table*}[!tph]
        \vskip 0.15in
        \centering
        \begin{tabular}{ccccccccc}
        \cline{3-8}
        & \multicolumn{1}{l|}{}  & \multicolumn{3}{c|}{\bf ROC-AUC}                                 &  \multicolumn{3}{c|}{\bf PR-AUC}                          &  \\ 
        \cline{1-8}
        \multicolumn{1}{|c|}{Dataset} & \multicolumn{1}{c|}{Model} & \multicolumn{1}{c|}{unseen} & \multicolumn{1}{c|}{seen} & \multicolumn{1}{c|}{mean} & \multicolumn{1}{c|}{unseen} & \multicolumn{1}{c|}{seen} & \multicolumn{1}{c|}{mean} & \\ 
        \cline{1-8}
        \multicolumn{1}{|c|}{CU-Bird200-2010} & \specialcell{ DA (baseline feat.) \cite{writeaclassifier} \\ DA+GP \cite{writeaclassifier} (baseline feat.)   \\ {DA \cite{whatyousee} (VGG feat.)} \\ Ours (fc baseline feat.) \\  Ours (fc) \\ Ours (conv) \\ Ours (fc+conv) }
        & \specialcell{0.59 \\ 0.62 \\ {0.66} \\ 0.69  \\ \bf 0.82 \\ 0.73 \\ 0.80  }
        & \specialcell{--- \\ --- \\ {0.69} \\ 0.93  \\ 0.96 \\ 0.96 \\ \bf 0.987  }
        & \specialcell{--- \\ --- \\ {0.68} \\ 0.85 \\ 0.934 \\ 0.91 \\ \bf 0.95 }
        & \specialcell{ --- \\ ---  \\ {0.037} \\ 0.09  \\ \bf 0.10 \\ 0.043 \\ 0.08 }
        & \specialcell{ --- \\ ---  \\ {0.11} \\ 0.20  \\ 0.41 \\ 0.34 \\ \bf 0.53 }
        & \multicolumn{1}{c|}{\specialcell{ --- \\ --- \\ {0.094} \\  0.19 \\ 0.35 \\ 0.28 \\ \bf 0.43 }} \ts\\
       \cline{1-8}
       \multicolumn{1}{|c|}{CU-Bird200-2011} & \specialcell{ Ours (fc) \\ Ours (conv) \\ Ours (fc+conv) }
        & \specialcell{ 0.82 \\ 0.80 \\\bf 0.85  }
        & \specialcell{ 0.974 \\ 0.96 \\ \bf 0.98  }
        & \specialcell{ 0.943 \\ 0.925 \\ \bf 0.953 }
        & \specialcell{ 0.11 \\ 0.085 \\\bf 0.13 }
        & \specialcell{ 0.33 \\ 0.15 \\ \bf 0.37 }
        & \multicolumn{1}{c|}{\specialcell{ 0.286 \\ 0.14 \\ \bf 0.31 }} \ts\\
       \cline{1-8}
       \multicolumn{1}{|c|}{Oxford Flower} & \specialcell{ DA (baseline feat.) \cite{writeaclassifier} \\ GPR+DA (baseline feat.) \cite{writeaclassifier} \\ Ours (fc baseline feat.) \\  Ours (fc) \\ Ours (conv) \\ Ours (fc+conv) }
        & \specialcell{0.62 \\ 0.68 \\  0.63  \\ 0.70 \\ 0.65 \\ \bf 0.71  }
        & \specialcell{--- \\ --- \\  0.96  \\ 0.987 \\ 0.97 \\ \bf 0.989  }
        & \specialcell{--- \\ --- \\ 0.86 \\ 0.90 \\ 0.85 \\ \bf 0.93 }
        & \specialcell{ --- \\ ---  \\ 0.055  \\ \bf 0.07 \\ 0.054 \\ 0.067 }
        & \specialcell{ --- \\ ---  \\ 0.60 \\ 0.65 \\ 0.61 \\ \bf 0.69 }
        & \multicolumn{1}{c|}{\specialcell{ --- \\ ---  \\ 0.45 \\ 0.52 \\ 0.46 \\ \bf 0.56 }} \ts\\
\cline{1-8}
\end{tabular}
\vskip 0.1in
\caption{\label{table:results}
ROC-AUC and PR-AUC(AP) performance compared to other methods. The performance is shown for both the zero-shot unseen classes and test data of the seen training classes. The class averaged mean AUCs are also included. For both ROC-AUC and PR-AUC, we report the best numbers obtained among the models trained on different objective functions.}
\vspace{-0.1in}
\end{table*}

In this section we empirically evaluate our proposed models and various objective functions. The $fc$ ({\it fully-connected}) model corresponds to Sec.~(\ref{sec:predfc}) where the text features are used to predict the fully-connected output weights of the image classifier. The $conv$ model is the convolutional classifier in Sec.~(\ref{sec:predconv}) that 
predicts the convolutional filters for CNN feature maps. The joint model is denoted as $fc$+$conv$. We evaluate the predicted zero-shot binary classifier on test images from both unseen and seen classes. The evaluation for Zero-Shot Learning performance varies widely throughout the literature. We report our model performance using the most common metrics:

\textbf{ROC-AUC:}  This is one of the most commonly used metrics for binary classification. We compute the receiver operating characteristic (ROC) curve of our predicted binary classifier and evaluate the area under the ROC curve. 

\textbf{PR-AUC(AP):}  It has been pointed out in \cite{auc} that for the dataset where the number of positive and negative samples are imbalanced, the precision-recall curve has shown to be a better metric compared to ROC. PR-AUC is computed by trapezoidal integral for the area under the PR curve. PR-AUC is also called average precision (AP). 

\textbf{Top-K classification accuracy:} Although all of our models can be viewed as binary classifiers, one for each class, the multi-class classification accuracy can be computed by evaluating the given test image on text descriptions from all classes and sorting the final prediction score $\hat{y}_c$.

\subsection{Training Procedure}

In all of our experiments, image features are extracted by running
the 19 layer VGG \cite{vgg} model pre-trained on ImageNet without
fine-tuning.  Specifically, to create the image features for the fully
connected classifier, we used the activations from the last fully connected 4096
dimension hidden layer fc1. The convolutional features are generated using 512x14x14
feature maps from the conv5\_3 layers. In addition, images are
preprocessed similar to~\cite{vgg} before being fed into the VGG net.
In particular, each image is resized so that the shortest dimension stays at 224
pixels. A center patch of 224x224 is then cropped from the resized image. 

Various components of our models are parameterized by ReLU neural nets of different sizes. The transformation function for textual features $f_t(\cdot):\mathbb{R}^d\mapsto \mathbb{R}^k$ are parameterized by a two-hidden layer fully-connected neural network whose architecture is p-300-k, where p is the dimensionality of the text feature vectors and $k=50$ is the size of the predicted weight vector $w_c$ for the fully connected layer. The image features from the fc1 layer of the VGG net are fed into the visual mapping $g_v(\cdot)$. This architecture is 4096-300-k. The intermediate convlayer features $a \in \mathbb{R}^{M\times w\times h}$ from the intermediate conv layer are first transformed by a conv layer $g_v'(\cdot)$ with $K'$ filters of size $3\times 3$, where we set $K'=5$. The final $a'\in\mathbb{R}^{K'\times w \times h}$ from Eq. \ref{eq:conv} are convolved with $K'\times 3 \times 3$ filters predicted from the 300 unit hidden layer of $f_t(\cdot)$. 

Adam \cite{adam} is used to optimize our proposed models with minibatches of 200 images. We found that SGD does not work well for our proposed models. This is potentially due to the difference in magnitude between the sparse gradient of the text features and the dense gradients in the convolutional layers. This problem is avoided by using adaptive step sizes. 

Our model implementation is based on the open-source package Torch~\cite{torch}. 
The training time for the fully connected model is 1-2 hours on a GTX Titan, whereas the joint fc+conv model takes 4 hours to train.

\subsection{Caltech UCSD Bird}

The 200-category Caltech UCSD bird dataset~\cite{cubird} is one of the most widely used and
competitive fine-grained classification benchmarks. We evaluated our method on
both the CUB200-2010 and CUB200-2011 versions of the bird dataset.
Instead of using semantic parts and attributes as in the common
approaches for CUB200, we only used the raw images and Wikipedia articles~\cite{writeaclassifier}
to train our models.  

There is one Wikipedia article associated with each bird class and 200
articles in total. The average number of words in the articles is around 400.
Each Wikipedia article is transformed into a 9763-dimensional Term Frequency-Inverse Document Frequency(tf-idf) feature vector. We noticed that
Log normalization for the term frequency is helpful, as article length
varies substantially across classes.

The CUB200-2010 contains 6033 images from 200 different bird species. There are
around 30 images per class. We follow the same protocol as in~\cite{writeaclassifier} using a random split of 40 classes as unseen and
the rest 160 classes as seen. Among the seen classes, we further allocate 
20\% of the images for testing and 80\% of images for training. There are around 3600 training set and 2500
images for testing. 5-fold cross-validation is used to evaluate the performance.

In order to compare with the previously published results, we first evaluated our
model using image and text features from~\cite{writeaclassifier}.
Since there are no image features with spatial information, we
are only predicting the fully connected weights.  Visual features are first fed
into a two-hidden layer neural net with 300 and 50 hidden units in the first and second layers.
We used their processed text features to predict the
50 dimensional fully connected classifier weights with a two hidden layer neural
net. A baseline Domain Adaptation~\cite{whatyousee} method is also evaluated using the
features from the VGG fc1 layer. 

The CUB200-2011 is an updated version of CUB200-2010 where the number of images are increased to 11,788. The 200 bird classes are the same as the 2010 version, but with the number of training cases doubled for each class. We used the same experimental setup and Wikipedia articles as the 2010 version.

\subsection{Oxford Flower}

The Oxford Flower-102 dataset\cite{oxflower} contains 102 classes with a total of 8189 images. The flowers were chosen from common flower species in the United Kingdom. Each class contains around 40 to 260 images. We used the same raw text corpus as in~\cite{writeaclassifier}. The experimental setup is similar to CUB200 where 82 flower classes are used for training and 20 classes are used as unseen during testing. Similar to the CUB200-2010 dataset, we compared our method to the previously published results using the same visual and text features. 

\subsection{Overall results}
Our results on the Caltech UCSD Bird and Oxford Flower datasets, shown in Table (\ref{table:results}), dramatically improve upon the state-of-the-art for zero-shot learning. This demonstrates that our deep approach is capable of producing highly discriminative feature vectors based solely on natural language descriptions. We further find that predicting convolutional filters (conv) and a hybrid approach (fc+conv) further improves model performance.

\begin{table}[t!]
        \centering
        \begin{tabular}{cccc}
        \cline{1-4}
        \multicolumn{1}{|c|}{Metrics} & \multicolumn{1}{c|}{BCE} & \multicolumn{1}{c|}{Hinge} & \multicolumn{1}{c|}{Euclidean} \\ 
        \cline{1-4}
        \multicolumn{1}{|c|}{\specialcell{unseen ROC-AUC  \\ seen ROC-AUC \\ mean ROC-AUC }}
        & \specialcell{\bf 0.82 \\\bf 0.973 \\ \bf0.937   }
        & \specialcell{0.795 \\ 0.97 \\ 0.934   }
        & \multicolumn{1}{c|}{\specialcell{ 0.70 \\ 0.95 \\ 0.90  }} \ts\\
       \cline{1-4}
        \multicolumn{1}{|c|}{\specialcell{unseen PR-AUC  \\ seen PR-AUC \\ mean PR-AUC }}
        & \specialcell{ \bf 0.103 \\ 0.33 \\ 0.287  }
        & \specialcell{ 0.10 \\\bf 0.41 \\ \bf 0.35  }
        & \multicolumn{1}{c|}{\specialcell{ 0.076 \\ 0.37 \\ 0.31 }} \ts\\
       \cline{1-4}
        \multicolumn{1}{|c|}{\specialcell{unseen class acc.  \\ seen class acc. \\ mean class acc. }}
        & \specialcell{0.01 \\ 0.35 \\  0.17  }
        & \specialcell{0.006 \\\bf 0.43 \\\bf  0.205  }
        & \multicolumn{1}{c|}{\specialcell{\bf 0.12 \\ 0.263  \\ 0.19 }} \ts\\
        \cline{1-4}
        \multicolumn{1}{|c|}{\specialcell{unseen top-5 acc.  \\ seen top-5 acc. \\ mean top-5 acc. }}
        & \specialcell{0.176 \\ 0.58 \\  0.38  }
        & \specialcell{0.182 \\\bf 0.668 \\  0.41  }
        & \multicolumn{1}{c|}{\specialcell{\bf 0.428 \\ 0.45  \\\bf 0.44 }} \ts\\
\cline{1-4}
\end{tabular}
\vskip 0.1in
\caption{\label{table:objfunc} Model performance using various objective functions on CUB-200-2010 dataset. The numbers are reported by training the fully-connected models.}
\end{table}

\begin{figure*}
\centering
\includegraphics[scale=0.5]{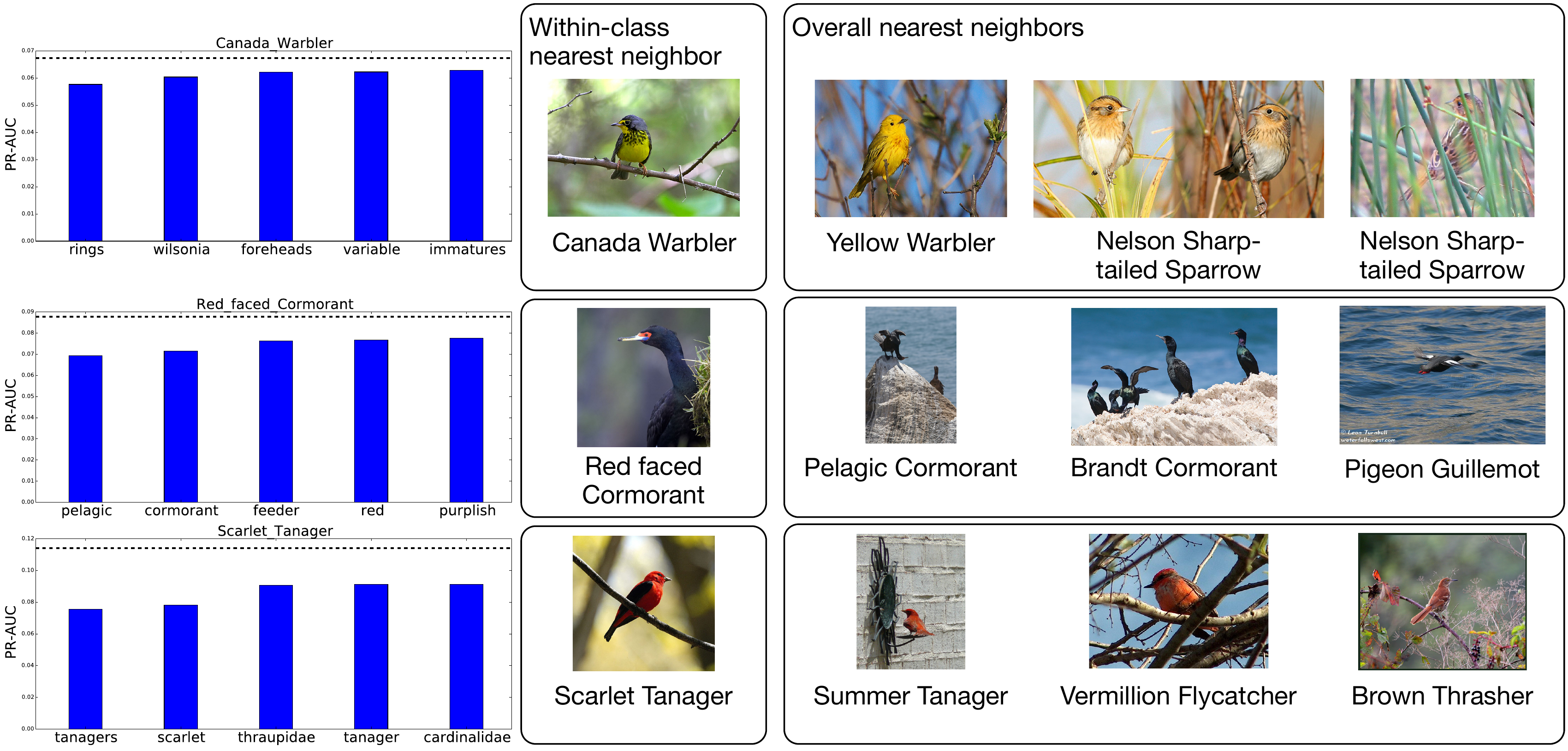}
\caption{[LEFT]: Word sensitivities of unseen classes using the fc model on CUB200-2010. The dashed lines correspond to the test-set PR-AUC for each class. TF-IDF entries are then independently set to 0 and the five words that most reduce the PR-AUC are shown in each bar chart. Approximately speaking, these words can be considered to be important attributes for these classes. [RIGHT]: The Wikipedia article for each class is projected onto its feature vector $w$ and the nearest image neighbors from the test-set (in terms of maximal dot product) are shown. The within-class nearest neighbors only consider images of the same class, while the overall nearest neighbors considers all test-set images.}
\label{fig:vis}
\vspace{-0.05in}
\end{figure*}

\subsection{Effect of objective functions}

We studied the model performance across the different objective functions from Sec.~\ref{sec:learning}. The evaluation is shown in Table (\ref{table:objfunc}). The models trained with binary cross entropy (BCE) have a good balance between ROC-AUC, PR-AUC and classification accuracy. The models trained with the hinge loss constantly outperform the others on the PR-AUC metric. However, the hinge loss models do not perform well on top-K classification accuracy on the zero-shot classes compared to other loss functions. The Euclidean distance model seems to perform well on the unseen classes while 
achieving a much lower accuracy on the seen classes. BCE shows the best overall performance across the three metrics.

\subsection{Effect of convolutional features}

The convolutional classifier and joint fc+conv model operate on the feature maps extracted from CNNs. Recent work~\cite{showattendtell} has shown that using features from convolutional layers is beneficial over just using the final fully connected layer features of a CNN. We evaluate the performance of our convolutional classifier using features from different intermediate convolutional layers in the VGG net and report the results in Table (\ref{table:convfeat}). The features from conv5\_3 layer are more discriminative than the lower Conv4\_3 layers. 
\begin{table}[t!]
        \centering
        \begin{tabular}{cccc}
        \cline{1-4}
        \multicolumn{1}{|c|}{Metrics} & \multicolumn{1}{c|}{Conv5\_3}& \multicolumn{1}{c|}{Conv4\_3} & \multicolumn{1}{c|}{Pool5} \\ 
        \cline{1-4}
        \multicolumn{1}{|c|}{\specialcell{mean ROC-AUC }}
        & \specialcell{\bf 0.91   }
        & \specialcell{ 0.6   }
        & \multicolumn{1}{c|}{\specialcell{ 0.82  }} \ts\\
       \cline{1-4}
        \multicolumn{1}{|c|}{\specialcell{mean PR-AUC }}
        & \specialcell{\bf 0.28  }
        & \specialcell{ 0.09  }
        & \multicolumn{1}{c|}{\specialcell{ 0.173 }} \ts\\
       \cline{1-4}
        \multicolumn{1}{|c|}{\specialcell{mean top-5 acc. }}
        & \specialcell{\bf 0.25 }
        & \specialcell{0.153  }
        & \multicolumn{1}{c|}{\specialcell{ 0.02 }} \ts\\
\cline{1-4}
\end{tabular}
\vskip 0.1in
\caption{ \label{table:convfeat}
 Performance comparison using different intermediate ConvLayers from VGG net on CUB-200-2010 dataset. The numbers are reported by training the joint fc+conv models.}
\end{table}

\subsection{Learning on the full datasets}

Similar to traditional classification models, our proposed method can be used for object recognition by training on the entire dataset. The results after fine-tuning are shown in Table (\ref{table:fullacc}).
\begin{table}[t1]
        \centering
        \begin{tabular}{cccc}
        \cline{1-4}
        \multicolumn{1}{|c|}{Model /\ Dataset} & \multicolumn{1}{c|}{CUB-2010} & \multicolumn{1}{c|}{CUB-2011} & \multicolumn{1}{c|}{OxFlower} \\ 
        \cline{1-4}
        \multicolumn{1}{|c|}{\specialcell{Ours (fc)  \\ Ours(fc+conv)  }}
        & \specialcell{0.60 \\\bf 0.62  }
        & \specialcell{0.64 \\\bf 0.66  }
        & \multicolumn{1}{c|}{\specialcell{ 0.73 \\ \bf 0.77  }} \ts\\
       \cline{1-4}
\end{tabular}
\vskip 0.1in
\caption{\label{table:fullacc}
  Performance of our model trained on the full dataset, a 50/50 split is used for each class.}
\end{table}

\subsection{Visualizing the learned attributes and text representations}
\label{sec:attribute}

Our proposed model learns to discriminate between unseen classes from text descriptions with no additional information. In contrast, more traditional zero-shot learning pipelines often involve a list of hand-engineered attributes. Here we assume that only text descriptions and images are given to our model. The goal is to generate a list of attributes for a particular class based on its text description.

Figure~\ref{fig:vis}, left panel, shows the sensitivity of three unseen classes on the CUB200-2010 test set using the fc model. For each word that appears in these articles, we set the corresponding tf-idf entry to 0 and measure the change in PR-AUC. We multiply by the ratio of the L2 norms of the tf-idf vectors before and after deletion to ensure that the network sees the same total input magnitude. The words that result in the largest decrease in PR-AUC are deemed to be the most important words (approximately speaking) for the unseen class.

In some cases the type of bird, such as ``tanager'', is an important feature. In other cases, physically descriptive words such as ``purplish'' are important. In other cases, non-descriptive words such as ``variable'' are found to be important, perhaps due to their rarity in the corpus. The collection of sensitive words can be thought of as pseudo-attributes for each class.

In Figure \ref{fig:vis}, right panel, we show the ability of the text features to describe visual features. For the three unseen classes, we use the text pipeline of the fc model to produce a set of weights, and then search the test set to find the images whose features have the highest dot product with the these weights. If we restrict the set of images to within the unseen class, we get the test image that is most highly correlated with its textual description. When we allow the images to span the entire set of classes, we see that the resulting images show birds that have very similar physical characteristics to the birds in the unseen classes. This implies that the text descriptions are informative of physical characteristics, and that the model is able produce a semantically meaningful joint embedding. More examples of these neighborhood queries can be found in the supplementary material.

\section{Limitations}

Although, our proposed method shows significant improvement on ROC-AUC over the previous method,  the multi-class recognition performance on the zero-shot classes, e.g. around 10\% top-1 accuracy on CUBird, is still lower than some of the attribute-based methods. It may be possible to take advantage of the discovered attribute list from Sec.~(\ref{sec:attribute}) to refine our classification performance. Namely, one may infer an attribute list for each class and learn a second stage attribute classification model. We leave this for future work.

\section{Conclusion}

We introduced a flexible Zero-Shot Learning model that learns to predict unseen image classes from encyclopedia articles. We used a deep neural network to map raw text and image pixels to a joint embedding space. This can be interpreted as using a natural language description to produce a set of  classifier weights for an object recognition network.

We further utilized the structure of the CNNs that incorporates both the intermediate convolutional feature maps and feature vector from the last fully-connected layer. We showed that our method significantly outperforms previous zero-shot methods on the ROC-AUC metric and substantially improves upon the current state-of-the-art on CUBird and Oxford Flower datasets using only raw images and text articles. We found that the network was able to learn pseudo-attributes from articles to describe different classes, and that the text embeddings captured useful semantic information in the images.

In future work, we plan to replace the tf-idf feature extraction with an LSTM recurrent neural network~\cite{hochreiter1997long}. These have been found to be effective models for learning representations from text.

\section*{Acknowledgments}
We gratefully acknowledge support from Samsung, NSERC and NVIDIA Corporation for the hardware donation.
{\small
\bibliographystyle{ieee}
\bibliography{0shot}

\begin{thebibliography}{10}\itemsep=-1pt

\bibitem{torch}
R.~Collobert, K.~Kavukcuoglu, and C.~Farabet.
\newblock Torch7: A matlab-like environment for machine learning.
\newblock In {\em BigLearn, NIPS Workshop}, number EPFL-CONF-192376, 2011.

\bibitem{auc}
J.~Davis and M.~Goadrich.
\newblock The relationship between precision-recall and roc curves.
\newblock In {\em ICML}, pages 233--240. ACM, 2006.

\bibitem{imagenet}
J.~Deng, W.~Dong, R.~Socher, L.-J. Li, K.~Li, and L.~Fei-Fei.
\newblock {ImageNet: A Large-Scale Hierarchical Image Database}.
\newblock In {\em CVPR}, 2009.

\bibitem{donahue2013decaf}
J.~Donahue, Y.~Jia, O.~Vinyals, J.~Hoffman, N.~Zhang, E.~Tzeng, and T.~Darrell.
\newblock Decaf: A deep convolutional activation feature for generic visual
  recognition.
\newblock {\em arXiv preprint arXiv:1310.1531}, 2013.

\bibitem{writeaclassifier}
M.~Elhoseiny, B.~Saleh, and A.~Elgammal.
\newblock {Write a Classifier: Zero-Shot Learning Using Purely Textual
  Descriptions}.
\newblock In {\em ICCV}, 2013.

\bibitem{farhadi2009describing}
A.~Farhadi, I.~Endres, D.~Hoiem, and D.~Forsyth.
\newblock Describing objects by their attributes.
\newblock In {\em CVPR}, pages 1778--1785. IEEE, 2009.

\bibitem{feifef03}
L.~Fe-Fei, R.~Fergus, and P.~Perona.
\newblock {A bayesian approach to unsupervised one-shot learning of object
  categories}.
\newblock In {\em CVPR}, 2003.

\bibitem{fink04}
M.~Fink.
\newblock {Object classification from a single example utilizing class
  relevance metrics}.
\newblock In {\em NIPS}, 2004.

\bibitem{frome2013devise}
A.~Frome, G.~S. Corrado, J.~Shlens, S.~Bengio, J.~Dean, T.~Mikolov, et~al.
\newblock Devise: A deep visual-semantic embedding model.
\newblock In {\em NIPS}, pages 2121--2129, 2013.

\bibitem{gopalan2011domain}
R.~Gopalan, R.~Li, and R.~Chellappa.
\newblock Domain adaptation for object recognition: An unsupervised approach.
\newblock In {\em ICCV}, pages 999--1006. IEEE, 2011.

\bibitem{hochreiter1997long}
S.~Hochreiter and J.~Schmidhuber.
\newblock Long short-term memory.
\newblock {\em Neural Computation}, 9(8):1735--1780, 1997.

\bibitem{khosla2012undoing}
A.~Khosla, T.~Zhou, T.~Malisiewicz, A.~A. Efros, and A.~Torralba.
\newblock Undoing the damage of dataset bias.
\newblock In {\em ECCV}, pages 158--171. Springer, 2012.

\bibitem{adam}
D.~Kingma and J.~Ba.
\newblock Adam: A method for stochastic optimization.
\newblock In {\em ICLR}, 2015.

\bibitem{alexnet}
A.~Krizhevsky, I.~Sutskever, and G.~E. Hinton.
\newblock Imagenet classification with deep convolutional neural networks.
\newblock In {\em NIPS}, pages 1097--1105, 2012.

\bibitem{whatyousee}
B.~Kulis, K.~Saenko, and T.~Darrell.
\newblock What you saw is not what you get: Domain adaptation using asymmetric
  kernel transforms.
\newblock In {\em CVPR}, pages 1785--1792. IEEE, 2011.

\bibitem{lampert09}
C.~H. Lampert, H.~Nickisch, and S.~Harmeling.
\newblock {Learning to detect unseen object classes by between class attribute
  transfer}.
\newblock In {\em CVPR}, 2009.

\bibitem{larochelle2008zero}
H.~Larochelle, D.~Erhan, and Y.~Bengio.
\newblock Zero-data learning of new tasks.
\newblock In {\em AAAI}, 2008.

\bibitem{mikolov2013distributed}
T.~Mikolov, I.~Sutskever, K.~Chen, G.~S. Corrado, and J.~Dean.
\newblock Distributed representations of words and phrases and their
  compositionality.
\newblock In {\em NIPS}, pages 3111--3119, 2013.

\bibitem{oxflower}
M.-E. Nilsback and A.~Zisserman.
\newblock Automated flower classification over a large number of classes.
\newblock In {\em ICVGIP}, pages 722--729. IEEE, 2008.

\bibitem{norouzi2014iclr}
M.~Norouzi, T.~Mikolov, S.~Bengio, Y.~Singer, J.~Shlens, A.~Frome, G.~S.
  Corrado, and J.~Dean.
\newblock Zero-shot learning by convex combination of semantic embeddings.
\newblock In {\em ICLR}, 2014.

\bibitem{Palatucci09}
M.~Palatucci, D.~Pomerleau, G.~E. Hinton, and T.~M. Mitchell.
\newblock {Zero-shot learning with semantic output codes}.
\newblock In {\em NIPS}, 2009.

\bibitem{parikh2011relative}
D.~Parikh and K.~Grauman.
\newblock Relative attributes.
\newblock In {\em ICCV}, pages 503--510. IEEE, 2011.

\bibitem{quionero2009dataset}
J.~Quionero-Candela, M.~Sugiyama, A.~Schwaighofer, and N.~D. Lawrence.
\newblock {\em Dataset shift in machine learning}.
\newblock The MIT Press, 2009.

\bibitem{saenko2010adapting}
K.~Saenko, B.~Kulis, M.~Fritz, and T.~Darrell.
\newblock Adapting visual category models to new domains.
\newblock In {\em ECCV}, pages 213--226. Springer, 2010.

\bibitem{vgg}
K.~Simonyan and A.~Zisserman.
\newblock Very deep convolutional networks for large-scale image recognition.
\newblock {\em arXiv preprint arXiv:1409.1556}, 2014.

\bibitem{socher2013zero}
R.~Socher, M.~Ganjoo, C.~D. Manning, and A.~Ng.
\newblock Zero-shot learning through cross-modal transfer.
\newblock In {\em NIPS}, pages 935--943, 2013.

\bibitem{torralba2011unbiased}
A.~Torralba and A.~A. Efros.
\newblock Unbiased look at dataset bias.
\newblock In {\em CVPR}, pages 1521--1528. IEEE, 2011.

\bibitem{cubird}
P.~Welinder, S.~Branson, T.~Mita, C.~Wah, F.~Schroff, S.~Belongie, and
  P.~Perona.
\newblock Caltech-ucsd birds 200.
\newblock 2010.

\bibitem{weston2011wsabie}
J.~Weston, S.~Bengio, and N.~Usunier.
\newblock Wsabie: Scaling up to large vocabulary image annotation.
\newblock In {\em IJCAI}, volume~11, pages 2764--2770, 2011.

\bibitem{showattendtell}
K.~Xu, J.~Ba, R.~Kiros, A.~Courville, R.~Salakhutdinov, R.~Zemel, and
  Y.~Bengio.
\newblock Show, attend and tell: Neural image caption generation with visual
  attention.
\newblock {\em arXiv preprint arXiv:1502.03044}, 2015.

\bibitem{yang2014unified}
Y.~Yang and T.~M. Hospedales.
\newblock A unified perspective on multi-domain and multi-task learning.
\newblock {\em arXiv preprint arXiv:1412.7489}, 2014.

\bibitem{deconv}
M.~D. Zeiler and R.~Fergus.
\newblock Visualizing and understanding convolutional networks.
\newblock In {\em ECCV}, pages 818--833. Springer, 2014.

\bibitem{ZhouNIPS2014}
B.~Zhou, A.~Lapedriza, J.~Xiao, A.~Torralba, and A.~Oliva.
\newblock {Learning Deep Features for Scene Recognition using Places Database}.
\newblock In {\em NIPS}, 2014.

\end{thebibliography}
}

\section{Appendix}

In the following figures, we first show more examples of the neighborhood queries for the fc models. We also visualize the predicted convolutional filters of conv models on both CU-bird and Oxford Flower datasets. The visualization of the predicted convolutional filter $w'_c$ from Section (3.3) are projected back to image space through the VGG 19 layer ImageNet model~\cite{vgg}. 

\begin{figure*}[ht]
\vskip 0.2in
\begin{center}
\centerline{\includegraphics[width=6.8in]{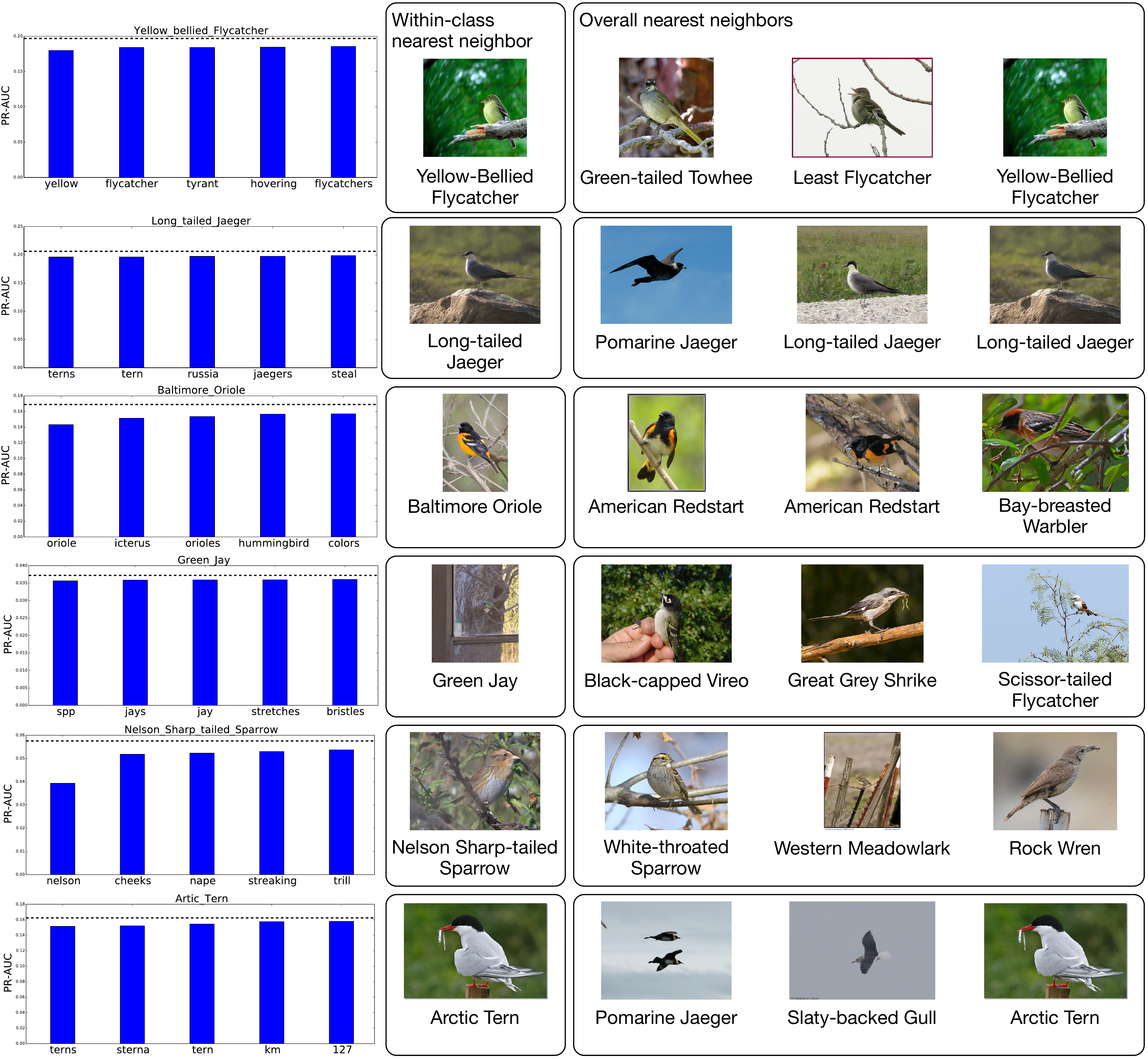}}
\caption{[LEFT]: Word sensitivities of unseen classes using the fc model on CUB200-2010. The dashed lines correspond to the test-set PR-AUC for each class. TF-IDF entries are then independently set to 0 and the five words that most reduce the PR-AUC are shown in each bar chart. Approximately speaking, these words can be considered to be important attributes for these classes. [RIGHT]: The Wikipedia article for each class is projected onto its feature vector $w$ and the nearest image neighbors from the test-set (in terms of maximal dot product) are shown. The within-class nearest neighbors only consider images of the same class, while the overall nearest neighbors considers all test-set images.}
\label{figure:attr}
\vskip -0.4in
\end{center}
\end{figure*} 

\begin{figure*}[ht]
\vskip 0.2in
\begin{center}
\centerline{\includegraphics[width=6.8in]{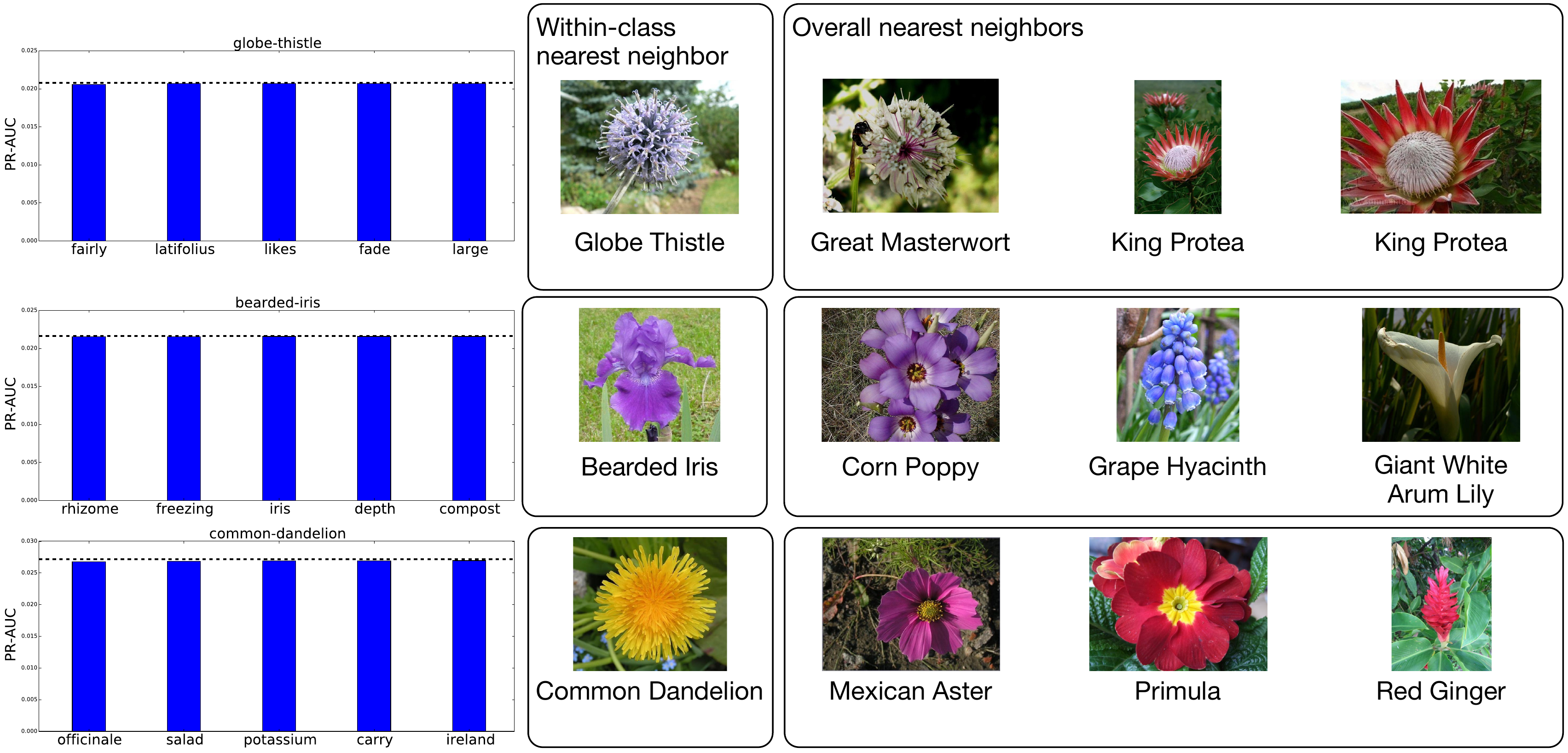}}
\caption{[LEFT]: Word sensitivities of unseen classes using the fc model on Oxford Flower. The dashed lines correspond to the test-set PR-AUC for each class. TF-IDF entries are then independently set to 0 and the five words that most reduce the PR-AUC are shown in each bar chart. Approximately speaking, these words can be considered to be important attributes for these classes. [RIGHT]: The Wikipedia article for each class is projected onto its feature vector $w$ and the nearest image neighbors from the test-set (in terms of maximal dot product) are shown. The within-class nearest neighbors only consider images of the same class, while the overall nearest neighbors considers all test-set images.}
\label{figure:attr}
\vskip -0.4in
\end{center}
\end{figure*}

\begin{figure*}[ht]
\vskip 0.2in
\begin{center}
\centerline{\includegraphics[width=6.8in]{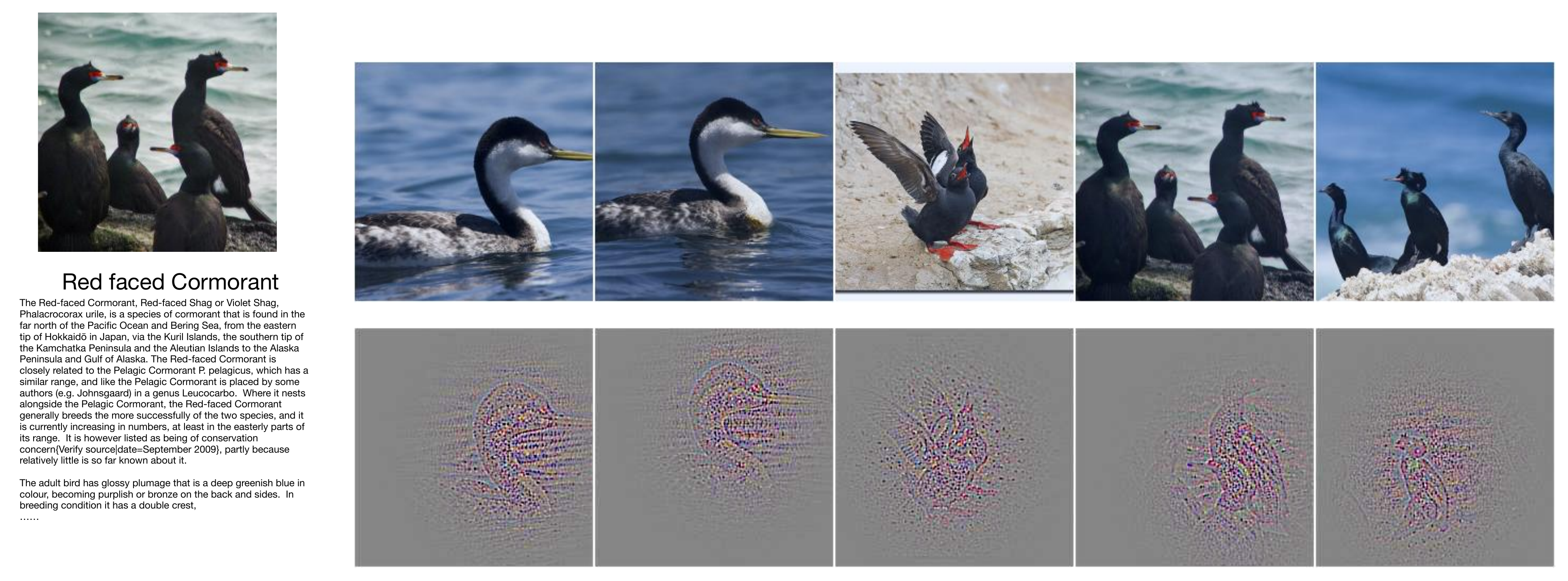}}
\centerline{\includegraphics[width=6.8in]{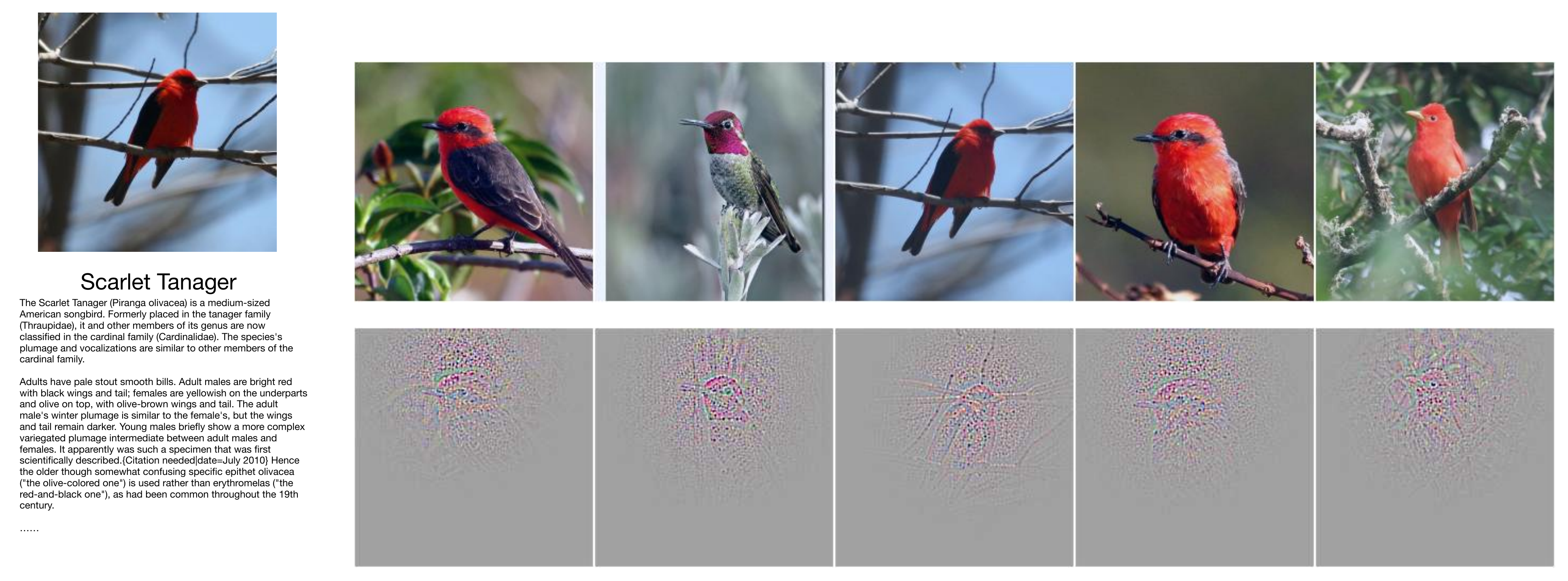}}
\centerline{\includegraphics[width=6.8in]{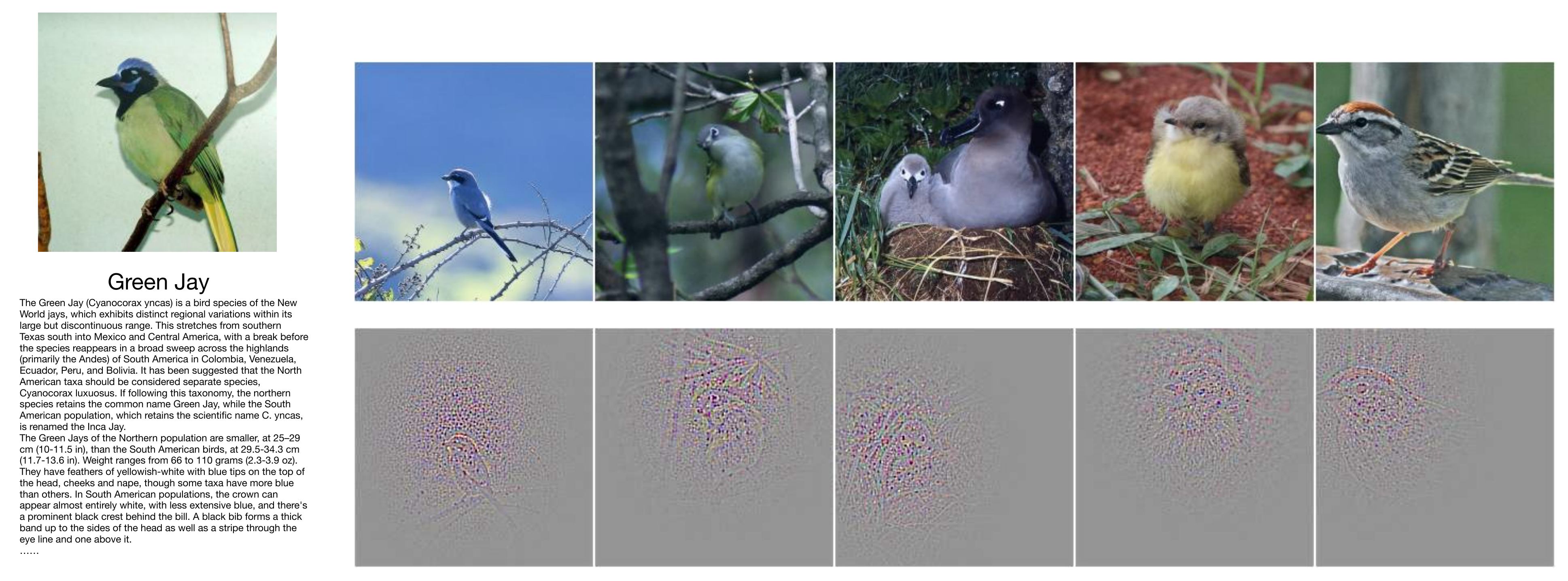}}
\caption{[LEFT]: Example images of unseen classes and their encyclopedia articles.  [RIGHT]: Visualizing the predicted convolutional filter $w'_c$ of the given unseen article on the left using the conv model trained on CUB200-2010. It shows top 5 images that have the highest activations for the predicted conv filters in the validation dataset (including both seen and unseen classes). The filter visualization under the images are generated by the deconvolution technique in \cite{deconv}. The highest activation in the predicted convolutional classifier is projected back into the image space. Best viewed in electronic version.}
\label{figure:attr}
\vskip -0.4in
\end{center}
\end{figure*} 

\begin{figure*}[ht]
\vskip 0.2in
\begin{center}
\centerline{\includegraphics[width=6.8in]{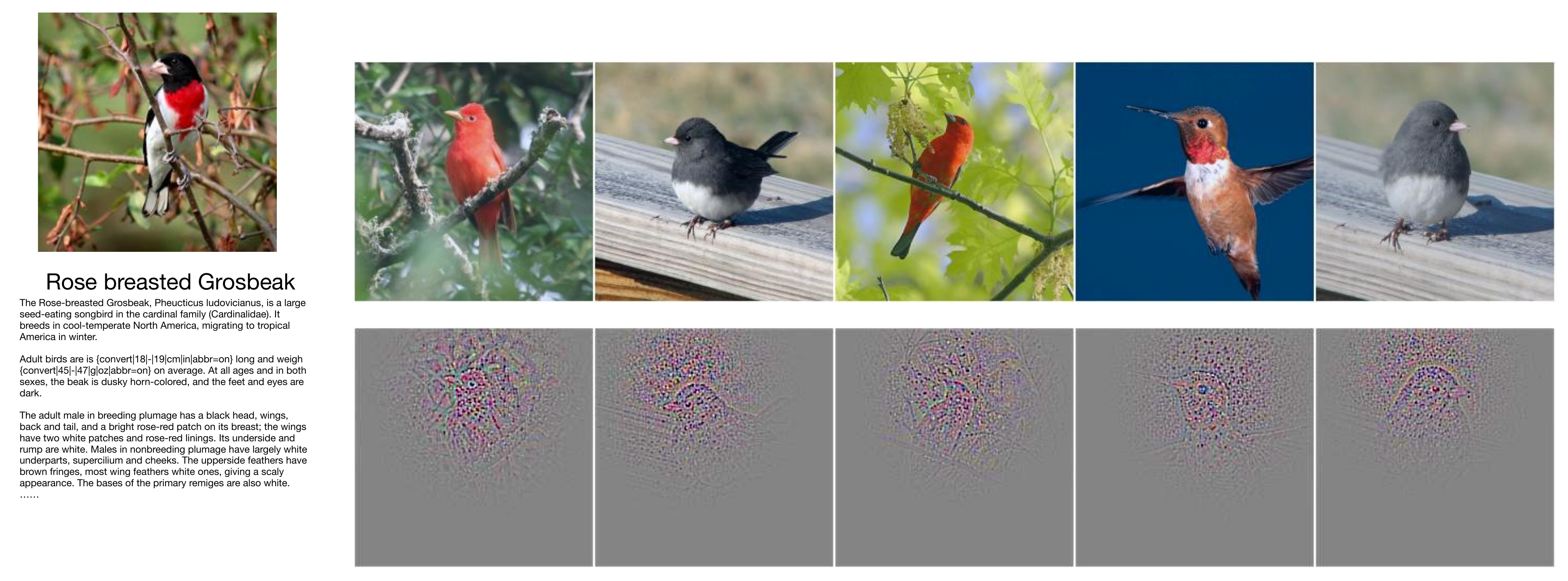}}
\centerline{\includegraphics[width=6.8in]{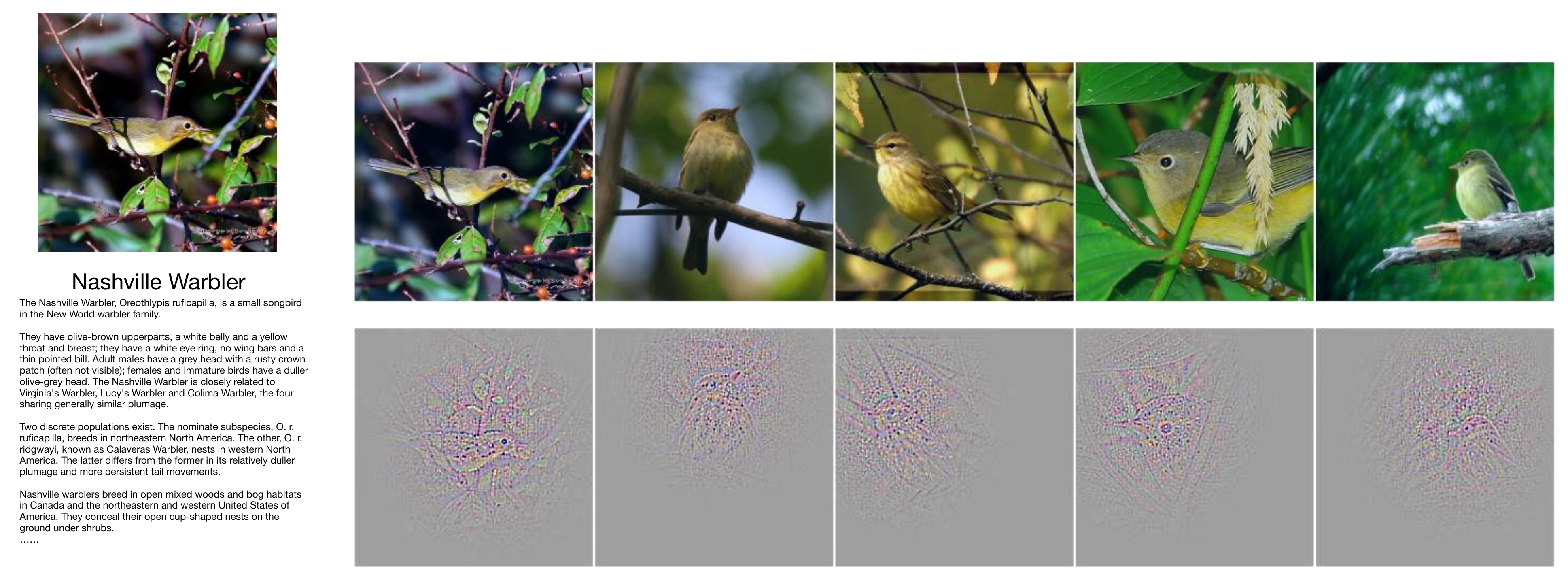}}
\centerline{\includegraphics[width=6.8in]{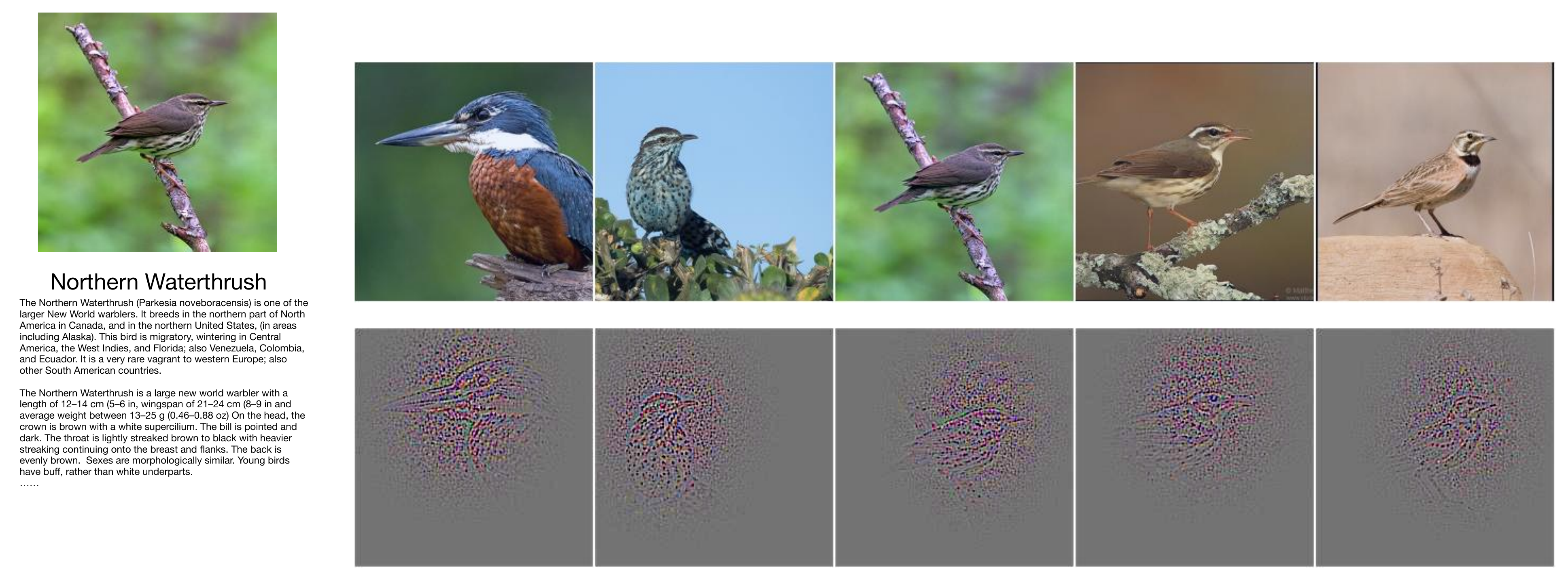}}
\caption{[LEFT]: Example images of unseen classes and their encyclopedia articles.  [RIGHT]: Visualizing the predicted convolutional filter $w'_c$ of the given unseen article on the left using the conv model trained on CUB200-2010. It shows top 5 images that have the highest activations for the predicted conv filters in the validation dataset (including both seen and unseen classes). The filter visualization under the images are generated by the deconvolution technique in \cite{deconv}. The highest activation in the predicted convolutional classifier is projected back into the image space. Best viewed in electronic version.}
\label{figure:attr}
\vskip -0.4in
\end{center}
\end{figure*} 

\begin{figure*}[ht]
\vskip 0.2in
\begin{center}
\centerline{\includegraphics[width=6.8in]{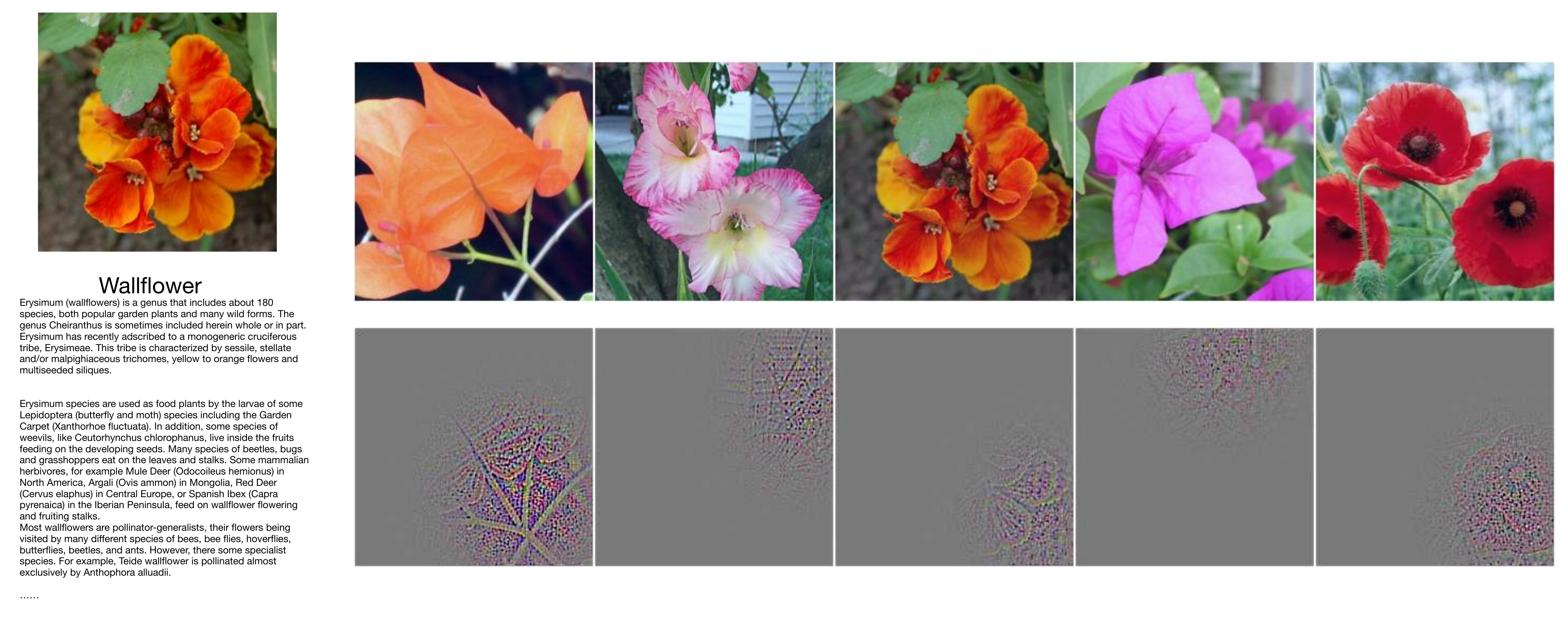}}
\centerline{\includegraphics[width=6.8in]{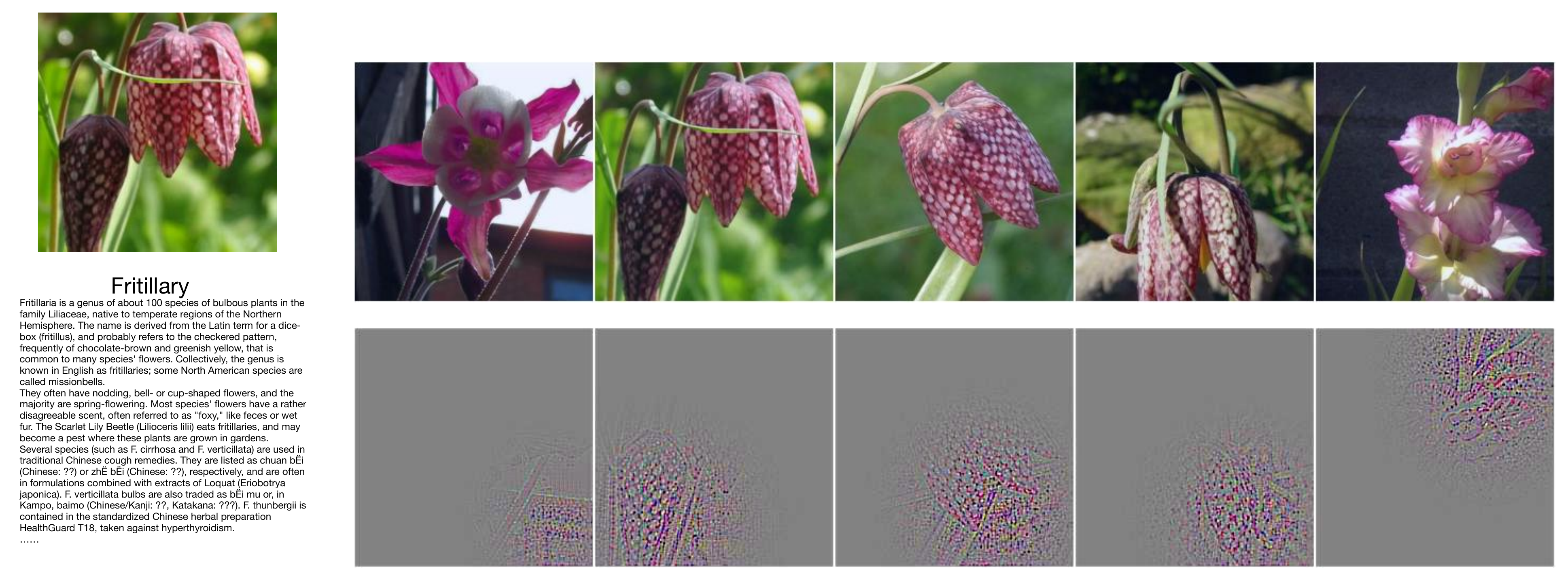}}
\centerline{\includegraphics[width=6.8in]{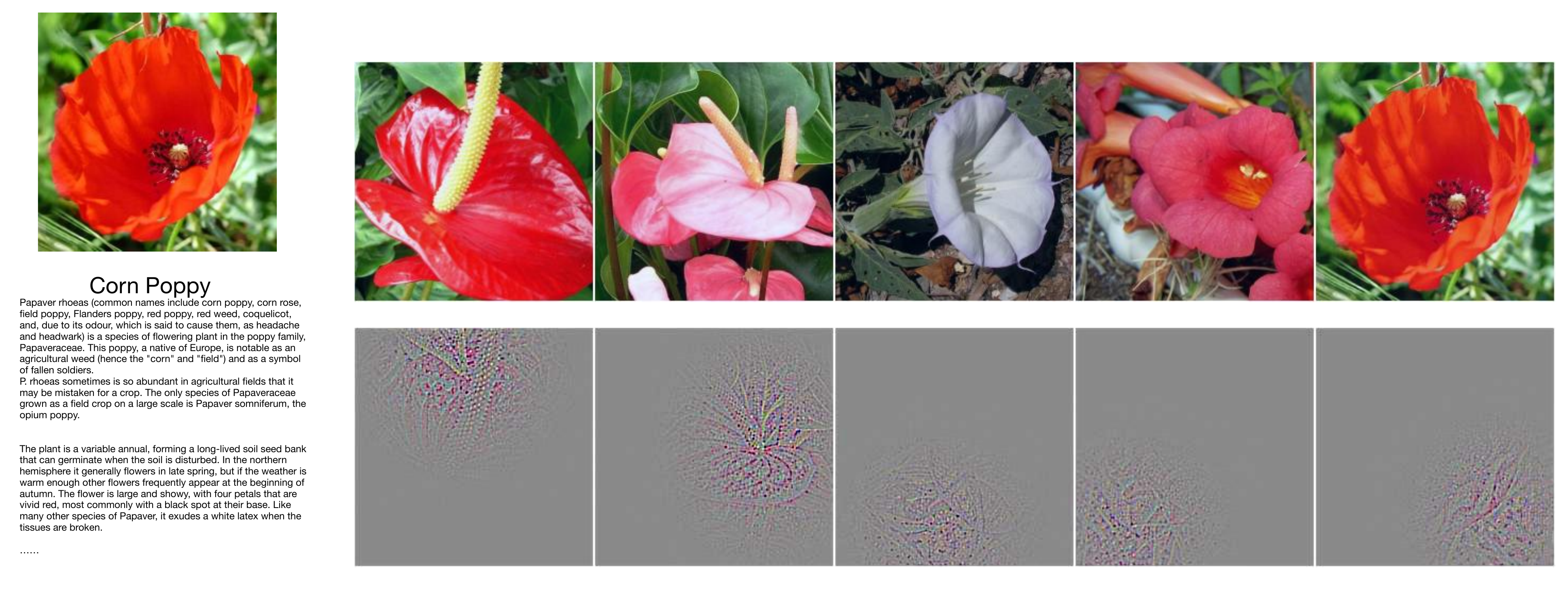}}
\caption{[LEFT]: Example images of unseen classes and their encyclopedia articles.  [RIGHT]: Visualizing the predicted convolutional filter $w'_c$ of the given unseen article on the left using the conv model trained on Oxford Flower. It shows top 5 images that have the highest activations for the predicted conv filters in the validation dataset (including both seen and unseen classes). The filter visualization under the images are generated by the deconvolution technique in \cite{deconv}. The highest activation in the predicted convolutional classifier is projected back into the image space. Best viewed in electronic version.}
\label{figure:attr}
\vskip -0.4in
\end{center}
\end{figure*} 

\begin{figure*}[ht]
\vskip 0.2in
\begin{center}
\centerline{\includegraphics[width=6.8in]{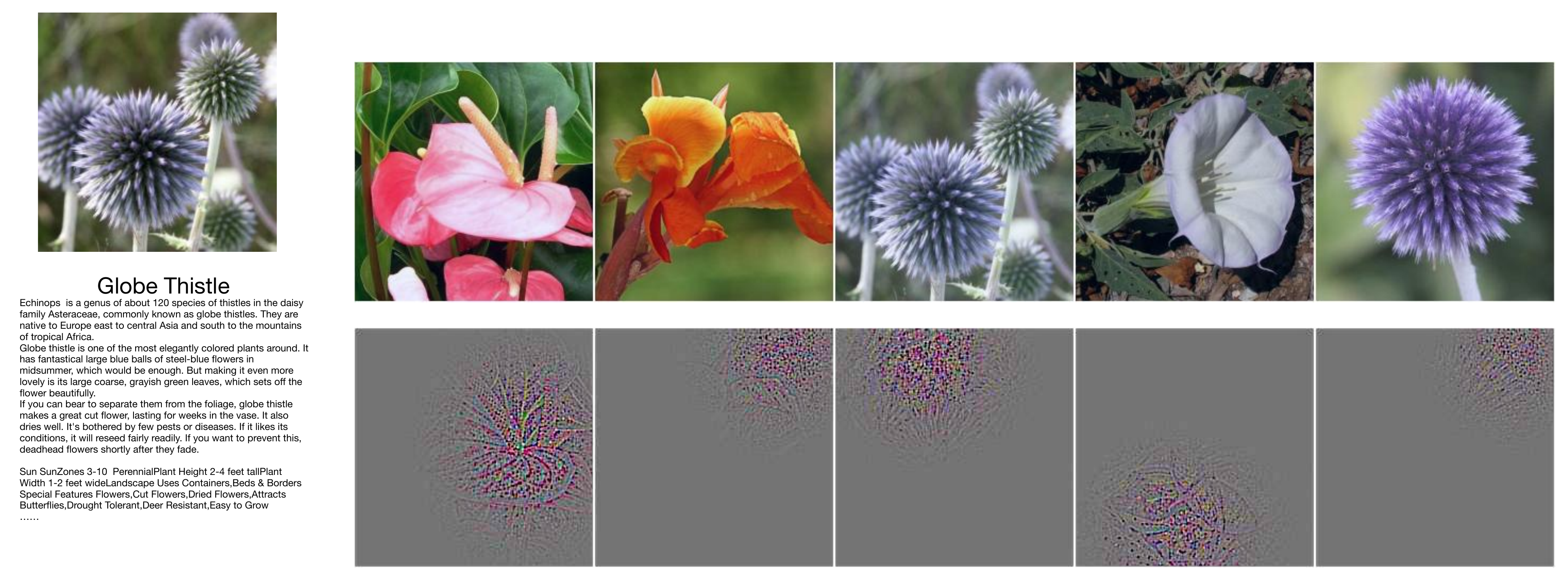}}
\centerline{\includegraphics[width=6.8in]{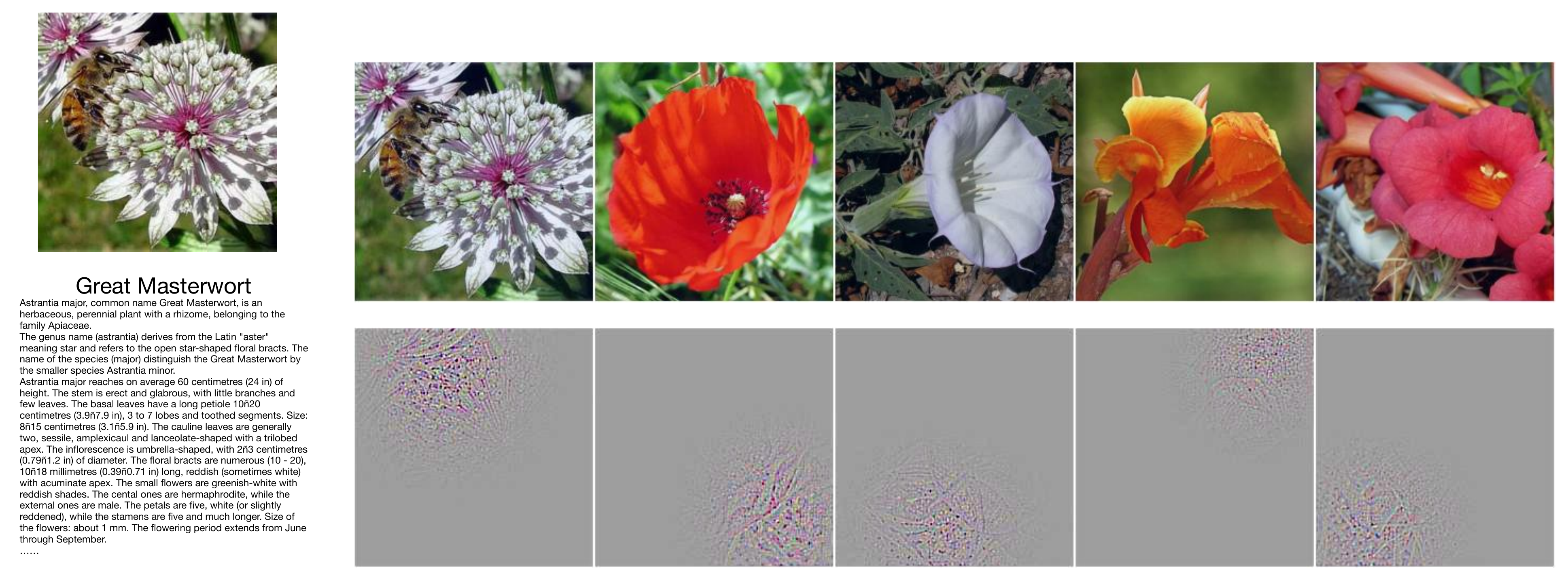}}
\centerline{\includegraphics[width=6.8in]{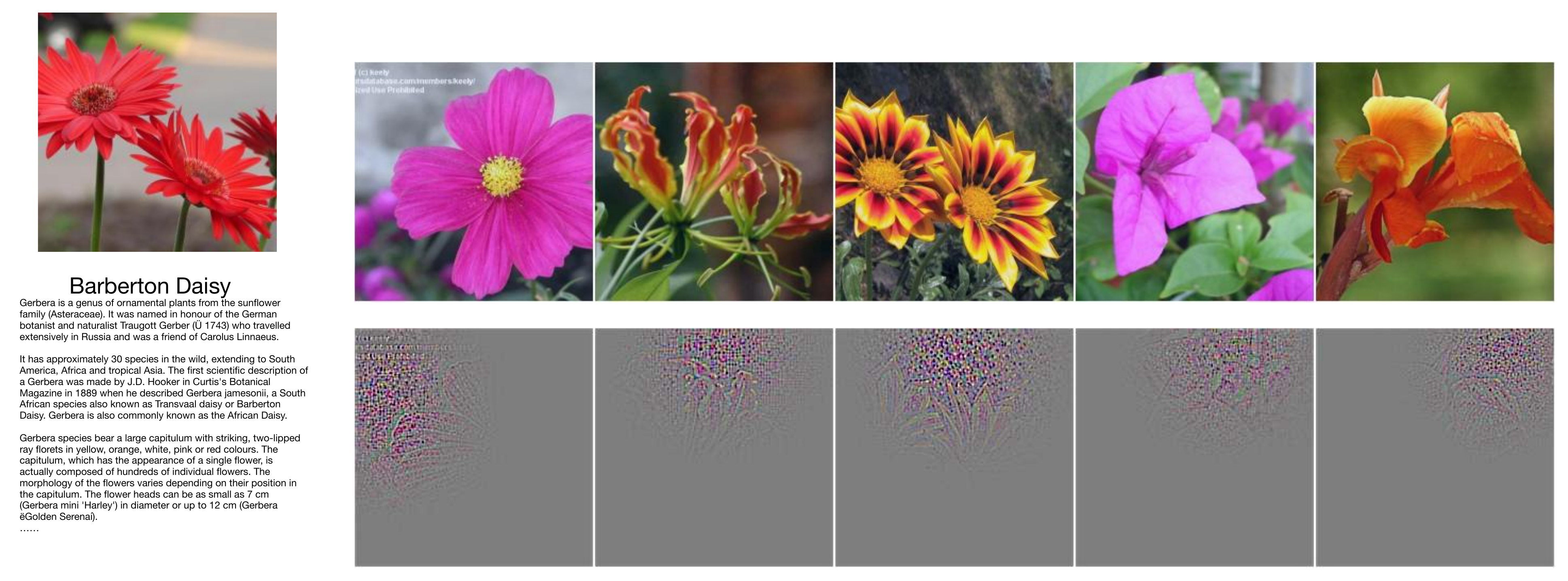}}
\caption{[LEFT]: Example images of unseen classes and their encyclopedia articles.  [RIGHT]: Visualizing the predicted convolutional filter $w'_c$ of the given unseen article on the left using the conv model trained on Oxford Flower. It shows top 5 images that have the highest activations for the predicted conv filters in the validation dataset (including both seen and unseen classes). The filter visualization under the images are generated by the deconvolution technique in \cite{deconv}. The highest activation in the predicted convolutional classifier is projected back into the image space. Best viewed in electronic version.}
\label{figure:attr}
\vskip -0.4in
\end{center}
\end{figure*}

\end{document}